\documentclass[journal, twocolumn ]{IEEEtran}
\ifCLASSINFOpdf
\else
\fi
\usepackage{graphicx}
\usepackage[table]{xcolor}
\usepackage{bm}
\graphicspath{{pictures/}}

\usepackage[colorinlistoftodos]{todonotes}[2cm]
\usepackage{siunitx}
\usepackage{pdfpages}

\usepackage{pdflscape}
\usepackage[section]{placeins}
\usepackage{longtable}
\usepackage{booktabs}
\usepackage{bm}
\usepackage{enumitem}
\usepackage{subfigure}
\usepackage{multirow}
\usepackage[hidelinks]{hyperref}

\usepackage{pgfplots}
\usepgfplotslibrary{external}

\let\oldlongtable\longtable
\let\endoldlongtable\endlongtable
\renewenvironment{longtable}{\fontfamily{phv}{}\selectfont\oldlongtable} {
\endoldlongtable \fontfamily{\familydefault}\selectfont}

\newcounter{row}

\usepackage{amsmath}
\usepackage{amssymb}

\usepackage{scalerel}
\usepackage{tikz}

\begin{document}

\title{Model-Based Capacitive Touch Sensing in Soft Robotics: Achieving Robust Tactile Interactions for Artistic Applications}

\author{Carolina Silva-Plata$^{\dagger\diamond}$, Carlos Rosel$^{\dagger}$, Barnabas Gavin Cangan$^{\ddagger}$, Hosam Alagi$^{\mathsection, \pm}$,  Björn Hein$^{\mathsection}$, \\Robert K. Katzschmann$^{\ddagger}$, Rubén Fernández$^{\diamond}$,Yosra Mojtahedi$^{\dagger\dagger}$ and Stefan Escaida Navarro$^{*\dagger}$

\thanks{$^{*}$Is corresponding author.$^{\dagger}$Authors are with the Instituto de Ciencias de la Ingeniería, Universidad de O'Higgins, $^{\diamond}$ Are with Departamento de Mecánica de la Universidad de Chile $^{\ddagger}$ Are with the Soft Robotics Lab (SRL) at ETH Zurich. $^{\mathsection}$ Are with the University of Applied Sciences Karlsruhe, Institut für Robotik und autonome Systeme. $^{\pm}$ Are with Proxception GmbH. $^{\dagger\dagger}$ is an independent artist. This work has been funded by the Agencia Nacional de Investigación y Desarrollo (ANID) in Chile with the Fondecyt the Iniciación grant number 11230505.}
}
\maketitle

\begin{abstract}
In this paper, we present a touch technology to achieve tactile interactivity for human-robot interaction (HRI) in soft robotics. By combining a capacitive touch sensor with an online solid mechanics simulation provided by the SOFA framework, contact detection is achieved for arbitrary shapes. Furthermore, the implementation of the capacitive touch technology presented here is selectively sensitive to human touch (conductive objects), while it is largely unaffected by the deformations created by the pneumatic actuation of our soft robot. Multi-touch interactions are also possible. We evaluated our approach with an organic soft robotics sculpture that was created by a visual artist. In particular, we evaluate that the touch localization capabilities are robust under the deformation of the device. We discuss the potential this approach has for the arts and entertainment as well as other domains.

\end{abstract}

\IEEEpeerreviewmaketitle

\section{Introduction}
\label{sec:Introduction}

\begin{figure}
\centering
\includegraphics[width=0.8\linewidth]{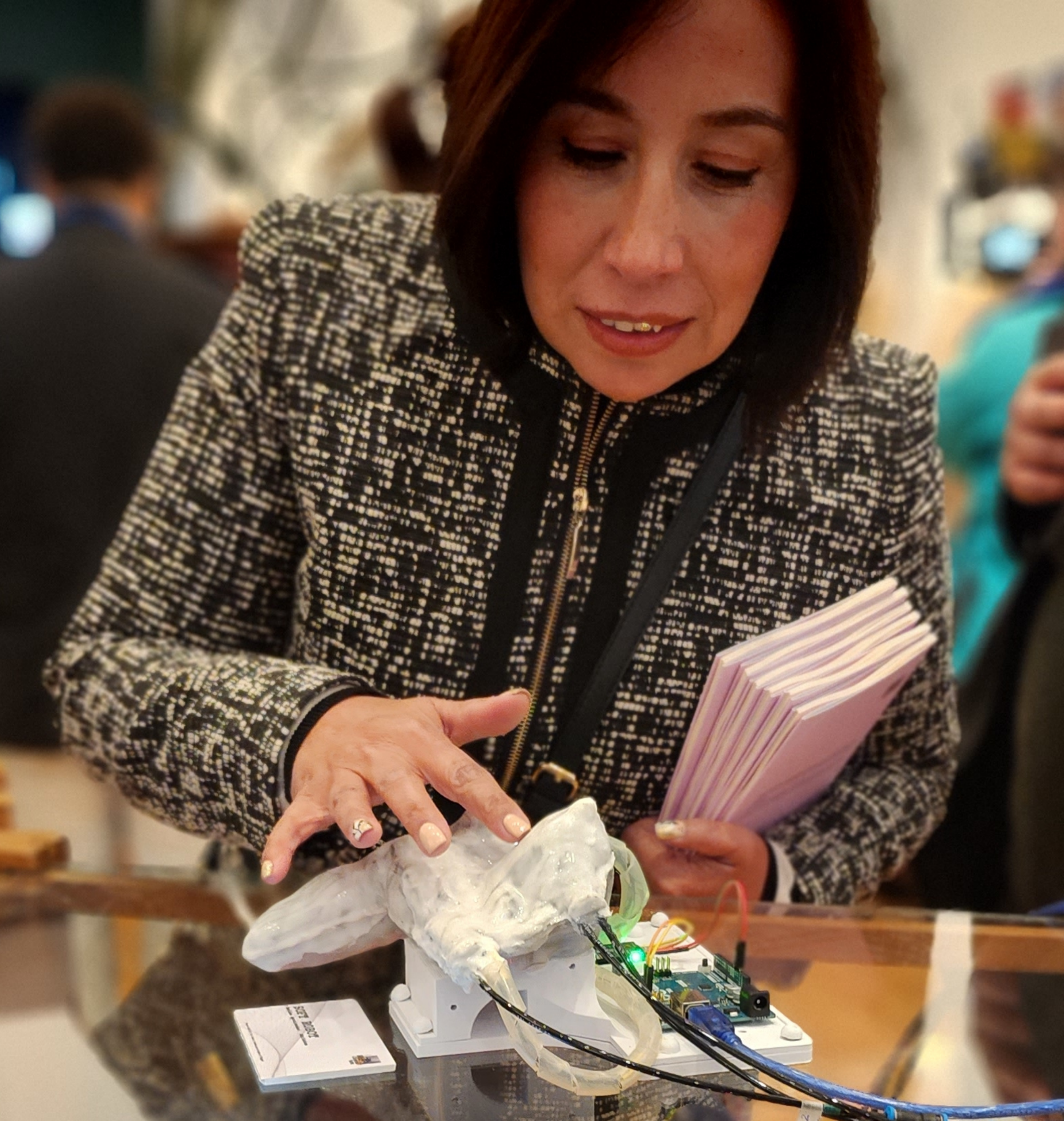}
\caption{A visitor at the museum uses touch to explore the interactive, sensorized soft robotics piece discussed in this work.}
\label{fig:Intro}
\end{figure}

Touch is an essential sensing modality for humans and other animals. For interactions with the environment, it delivers vital cues about the material properties of objects, such as compliance, texture and temperature. In the context of social interactions, it conveys a variety of cues that enable, for instance, effective collaboration through force exchange or display of intimacy and affection. An effective touch technology for soft robotics can unlock a variety of novel applications in domains such as the arts/entertainment, medicine, agriculture, etc. The technology we propose in this paper is mainly demonstrated by a novel visual/tactile soft robotics sculpture called ``El Fruto de L'Érosarbénus''.\footnote{This piece is derivated from a previous soft robotics sculptural installation ``L'Érosarbénus'' (2020) by the collaborating artist Yosra Mojtahedi.} This piece, which has touch as its leading interaction modality, was shown at an exhibition in Rancagua, Chile, during May 2024. Fig.~\ref{fig:Intro} shows a visitor of the exhibition interacting with the piece. We call the piece \emph{The Fruit} in this work. 
 
In robotics, tactile and proximity skins have been developed since the early days of the field to address challenges related to human-robot interaction (HRI), collision avoidance, and object manipulation~\cite{navarro2021proximity}. Important advances have been made particularly in classical rigid robotics, where large body coverage of artificial skins is state of the art. When designing a tactile skin for a rigid robot link, deformation is typically exploited to transduce the location and force of the interaction. For example, using capacitive sensing, the change in distance between electrodes separated by a compressible and insulating layer can be used to achieve such a measurement of force. When the principle is replicated multiple times over a given surface, spatial resolution is achieved and, with it, the ability to locate touch. However, in soft robotics, exploiting deformation for sensing has important inconveniences. This is because a soft robot achieves its change in configuration by inducing deformation of a part of the robot or the robot as a whole. Thus, a sensing principle for touch that relies on transducing deformations faces the dilemma of how to distinguish in the signals what the contributions are due to a change in configuration and what contributions are due to external interactions, since both are bound to be reflected in the sensor signal.

In this work, we study an implementation of capacitive touch technology for HRI, which is largely unaffected by the deformations created by the actuation of the soft robot. In our case, the soft robotics sculpture is actuated pneumatically, implementing an animation. The proposed technology also allows multitouch. This work is based on our recent results regarding model-based sensing~\cite{navarro2020model,cangan2022model,tian2023multi} where we address the challenge of tactile and shape sensing in soft robotics. In our approaches for model-based sensing, a solid-mechanics simulation based on the Finite Element Method (FEM) implemented in SOFA framework (Simulation Open Framework Architecture)~\cite{coevoet2017software}, helps to interpret the sensor signals observed.  In particular, the simulation represents the employed sensing modality and models the deformation caused by actuation principles, such as pressurized air or tendons. It is important to note that this work focuses on touch \emph{localization} with actuation modeling. Estimating the force and deformation resulting from touch is a logical extension but was not feasible within the constraints of the proposed art piece.

We make the following contributions in this work:
\begin{enumerate}
\item We show a capacitive tactile technology for HRI that remains largely unaffected by actuation-induced deformation. It can reliably detect contact for the robot, even in its deformed state.
\item We show that this skin can conform to arbitrary surfaces making it well suited for novel applications in the arts, entertainment, or medical fields.
\item We show that this skin has multitouch capabilities.
\end{enumerate}
For a quick overview of these contributions, we suggest that the reader refers to the video accompanying this paper. The remainder of the paper is structured as follows: In the next section, we review related work in the field. In Sec.~\ref{sec:CTT}, we discuss the capacitive touch technology used. Then, in Sec.~\ref{sec:Modeling}, we review the mechanical model employed for simulation and how it can be used to interpret touch. In Sec.~\ref{sec:Results}, we discuss our experimental validation. Finally, in Sec.~\ref{sec:Conclusions}, we provide our conclusions for these contributions and give an outlook for future research directions.
\section{Related Work}
\label{sec:RelatedWork}
In~\cite{teyssier2019skin}, Teyssier et al.\ introduced a novel method to extend mutual capacitance-based touch detection to soft bodies using a stretchable conductive yarn called \emph{DataStretch} and a capacitive multi-touch controller by \emph{FocalTech} deployed on a breakout board, featuring a resolution of up to $21\!\times\!12$ taxels. This enabled the integration of human-like artificial skin sensors in modified consumer electronic devices that can precisely detect touch. Parilusyan et al.~\cite{parilusyan2022sensurfaces} extended this further and developed a modular hardware platform that allowed the implementation of sensing surfaces on a wide variety of materials. They demonstrated this capability in applications such as sensitive concrete floors, sensitive soft furnishings, seamless smart home, etc. 

Dawood et al.~\cite{dawood2023icra} presented a soft e-skin based on capacitive electrodes embedded between layers of silicone. Using Machine Learning (ML) approaches such as simple Linear Regression, Random Forest Classifiers and Gaussian Process Regression, they were able to demonstrate remarkable accuracy in force estimation with multiple contacts. However, in their multi-touch experiments, the two contacts needed to be spaced apart by at least 4 nodes. This is because they used a classification approach with densely positioned flat conductors that are rather wide relative to the spacing. 

In~\cite{truby2018soft}, Truby et al. present a multi-modal sensor design for a soft finger that is able to detect bending, actuation and touch independently. The modalities are decoupled by a sophisticated design, fabricated with specialized 3D printing techniques. However, touch interaction is only detected at single point.

Lampinen et al.~\cite{lampinen2024soft} proposed an approach to multi-axis soft force sensing with a pneumatic tactile pressure sensor matrix operating on fluidic principles. The sensors, made of elastomers, contain embedded microchannels and forces were measured by sensing changes in pressure with deformation. By using multiple sensors in a matrix, they showed the ability to sense not only magnitude of the contact forces but also the location of contact. Since the microchannels are fabricated to have a specific shape, they were able to achieve this without requiring extensive modeling of the sensor. However, the authors do not study the effect of underlying deformation of a soft robot on the sensor signals.

Kim et al.~\cite{kim2024multi} built a novel modular textile sensor array and developed an algorithm for multi-modal sensing of human touches with contact forces. Their sensor is fabricated by simple sequential lamination of conductive and non-conductive materials. However, their approach addressed the typically challenging wiring problem by using connecting multiple capacitive sensing modules using only two wires. To achieve this, they use bandstop filtering and a compensation algorithm derived from frequency domain analysis of the transfer function of the sensor. This approach allowed them to connect about 13 sensors to only two wires.

In terms of model-based sensing techniques, in \cite{tapia2020makesense}, Tapia et al.\ present an approach for optimal sensorization of soft robots using stretch sensors to reconstruct their configuration. The robot's mechanics are modeled using the FEM and a model for the resistive type sensors is established, allowing to relate deformations to sensor values and vice versa. This allows finding the deformations due to actuation efforts and external forces. However, this approach lacks the possibility to localize the external contacts, and thus the authors ``[...] estimate contact forces of unknown magnitude and direction on the entire surface of the gripper.'' 

Overall, to the best of our knowledge, there is no comprehensive approach for detecting touch for HRI in soft robotics that provides: 1) robust contact detection under deformation for arbitrary shapes 2) A link of this detection to a interactive mechanical model in order provide generalized touch position estimation under actuation.

\section{Capacitive Touch Technology}
\label{sec:CTT}
\begin{figure}
    \centering
    \footnotesize
    \includegraphics[width=0.95\linewidth]{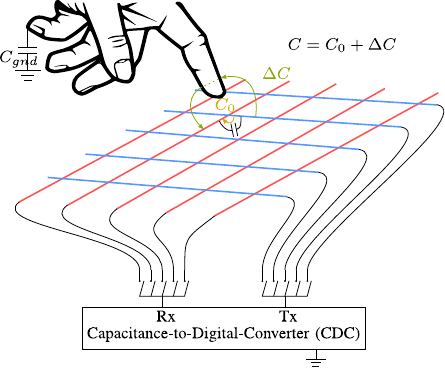}
    \caption{The total capacitance at each overlapping Tx and Rx is the sum of $C_{0}$ due to the initial coupling between Tx and Rx und the $\Delta C$ due to coupling through the touching human finger.}
    \label{fig:MuCaPrinciple}
\end{figure}

\subsection{Soft Mutual Capacitive Sensing}
\label{subsec:Muca}

\begin{figure}
    \centering
    \footnotesize
    \includegraphics[width=0.80\linewidth]{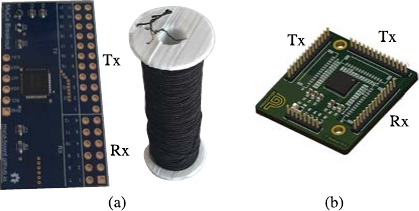}
    \caption{(a) The \emph{MuCa Kit}, proposed by Teyssier et al.~\cite{teyssier2019skin} (b) Our own version of the breakout board, used in this work.}
    \label{fig:Yarn+Electronics}
\end{figure}

In terms of electronics, we use the combination of capacitive-to-digital converter chip (CDC) by \emph{FocalTech} with a conductive yarn called \emph{DataStretch} that was first proposed for this type of applications by Teyssier et al.~\cite{teyssier2019skin} (see Fig.~\ref{fig:Yarn+Electronics}(a)). This project is called \emph{MuCa}. The DataStretch yarn's bending stiffness is negligible compared to the elastomer material and it can stretch up to \SI{30}{\percent}. To avoid the effect of the stretching stiffness, we propose a slight zig-zag sewing pattern when fabricating the skin. Due to the temporal unavailability of the CDC breakout, we decided to implement our own version of the board for this work (see Fig.~\ref{fig:Yarn+Electronics}(b)). To implement mutual capacitance touch measurements, crossing transmitter (Tx) and receiver (Rx) stripes create an interaction surface by multiplexing the Tx and Rx channels. $C_{0}$ represents the initial capacitive coupling at each crossing point. A conductive object in the nearby vicinity (practically touching) disturbs the baseline electric field and causes changes in the capacitance, here represented by $\Delta C$. The CDC captures the total capacitance $C=C_{0}+\Delta C$ and converts it into a digital value for each measurement point. As mentioned in Sec.~\ref{sec:RelatedWork}, the \emph{FocalTech} CDC can implement up to $21\!\times\!12$ touch points. For further details about capacitive sensing in HRI and HMI (Human-Machine Interaction), please refer to the following survey papers:~\cite{navarro2021proximity,grosse2017finding}. 

\label{subsec:Muca}

\subsection{Weighted Position Detection and Multitouch}
\label{subsubsec:WeightedPositionDetection}
\begin{figure}
    \centering
    \small
    \includegraphics[width=0.7\linewidth]{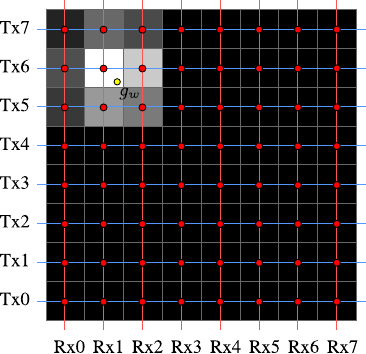}
    \caption{An example of an activation map on an $8\!\times\!8$ grid of taxels and a weighted touch position estimate $g_{w}$.}
    \label{fig:GridWeights}
\end{figure}

The intersection points between the Tx and Rx wires define individual sensing elements, referred to as taxels (tactile elements), which are centered at the grid points $g_{ij}$. These taxels act as discrete sensing locations, where the capacitance is measured to detect touch. As shown in Fig.~\ref{fig:GridWeights}, the grid of taxels forms a tactile image that can localize a touch event based on the intensity values  $v_{ij}$ observed at each intersection point. From this activation map we can deduce a \emph{weighted} touch position $g_{w}$ in the grid coordinate system. For this, we consider the 9-Neighborhood $N_{9}$ of the taxel having the maximal activation, i.\,e.\ its 8 neighbors plus the maximally activated taxel itself. Let $v_{sum}=\sum v_{ij}$, with $(i,j)\in N_{9}$, be sum of activation values of $N_{9}$. Then, each weight is:
\begin{equation}
    w_{ij} = \frac{v_{ij}}{v_{sum}}.
\end{equation}
The coordinates of the weighted points are then:
\begin{equation}
\label{eq:WeightedPosition}
    g_{w} = \sum_{(i,j)\in N_{9}}w_{ij}g_{ij}. 
\end{equation}
For multitouch, more than one touch event can be localized by determining maximum values sequentially. After one event has been processed like described above, the activation map is updated. All values inside $N_{9}$ are set to 0. If there are other taxels remaining with activations above the threshold, the new maximum is found, and the process for finding the weights in this new 9-Neighborhood is repeated. Later, in Sec.~\ref{subsec:Mapping2D_3D}, we discuss how to transfer the touch positions on the 2D grid to the arbitrary shape on the model.
The attached video shows an example of how our implementation of multitouch detection works. 

\subsection{Design for Robustness to Underlying Deformation}
\label{subsec:Robustness}
\begin{figure}
    \centering
    \includegraphics[width=0.9\linewidth]{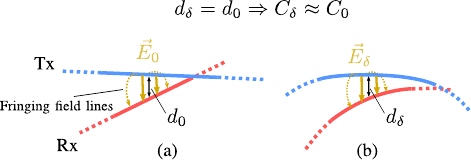}
    \caption{(a) Electric field configuration $\vec{E}_{0}$ for the taxel at rest. (b) Electric-field configuration $\vec{E}_{\delta}$ for the taxel experiencing the effects of underlying deformation. If the distance between the electrode layer does not change, the baseline capacitances $C_{0}$ and $C_{\delta}$ are expected to be very close.}
    \label{fig:RobustnessDeformation}
\end{figure}

The property of mutual capacitive sensing that enables robustness against underlying deformation $\delta$ is illustrated in Fig.~\ref{fig:RobustnessDeformation}. The electric field responsible for the baseline capacitance at rest $\vec{E}_{0}$ is concentrated around the crossing point of the electrodes (see Fig.~\ref{fig:RobustnessDeformation} (a)). For a constant voltage amplitude, its strength is mainly determined by the geometry of the conductors and distance $d_{0}$ between them. Only a small amount of capacitance is due to the \emph{fringing field lines}, which extend farther from the crossing point. 
To ensure robustness, we propose to design the interaction surface such that the electrodes are glued together at the crossing points, only separated by a thin insulating layer. As a result, the distance at these crossing point does not change when the underlying material deforms. Although deformation stretches or compresses the fringing field lines (see Fig.~\ref{fig:RobustnessDeformation} (b)), this only causes a slightly different baseline capacitance $C_{\delta}$. This difference is small compared to the change in capacitance $\Delta C$ due to an external approaching object, illustrated in Fig.~\ref{fig:MuCaPrinciple}. Thus, this design provides robustness with regards to underlying deformation.

\subsection{Integration Into Arbitrary Surfaces}
\label{subsec:Integration}
\begin{figure}
    \centering
    \small
    \includegraphics[width=0.99\linewidth]{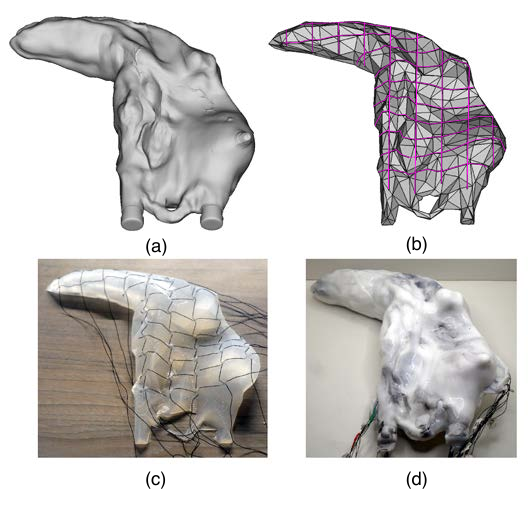}
    \caption{(a) The original high-res triangular mesh of the Fruit. (b) The reduced mesh with guiding tracks for the conductive yarns, created algorithmically. (c) The fabricated silicone piece that includes the conductive yarn in a slight zig-zag pattern. (d) The complete fabrication result of two halves glued together, with the covering layer and embedded chambers.}
    \label{fig:FabricationThreads}
\end{figure}

On the example of the Fruit, we illustrate how the integration can be achieved for arbitrary surfaces. The process explained in the following is illustrated by Fig.~\ref{fig:FabricationThreads} (a)-(d). We have implemented a grid with 47 points on the fruit. The vertical spacing is \SI{12}{\milli\meter} and horizontal spacing is \SI{16}{\milli\meter}. In order to be able to locate the contact on an arbitrary surface, it is essential to determine the 3D positions of the grid intersection points. Furthermore, it is useful for fabrication to have guides/tracks for the conductive yarn to follow. 
To achieve this, we generate a slightly smaller version of the top half of the original shape, with a significantly reduced triangle count. The smaller size leaves space for a covering layer that provides the desired appearance, while the reduced triangle count ensures computational efficiency and robustness during geometric operations.
On the reduced shape, we create the tracks using an algorithm implemented using the OCC (Open Cascade Technology) backend in \emph{Gmsh}~\cite{geuzaine2009gmsh}, specifically designed for this purpose. In the algorithm, the shape generated is cut with planes in the horizontal and vertical directions, corresponding to the desired tactile grid. Each cut produces a curve, and from each curve,  extrude pipe-like shapes are generated and subtracted from the reduced shape to create the tracks (see Fig.~\ref{fig:FabricationThreads}~(b)). This process also identifies the 3D intersection points of the tracks, corresponding to the taxel positions in 3D, which are later used to determine the touch coordinates in the simulation. Then, the reduced shape is fabricated in silicone by a casting process. We used Ecoflex 00-50 silicone (Smooth-On) for the entire piece. Using the guides provided by the tracks, the conductive yarn is sewn into the shape in a slight zig-zag pattern (see Fig.~\ref{fig:FabricationThreads}~(c)). A next step adds a covering layer to achieve the desired appearance. {Simultaneously,} the lower half of the shape is cast. The final Fruit results from gluing both halves together (see Fig.~\ref{fig:FabricationThreads}~(d)). 

\section{Mechanical Model in SOFA}
\label{sec:Modeling}
The SOFA framework implements the FEM, which yields the internal elastic forces $\mathbb{F}(\bm{q})$, given that the nodes of the FEM-mesh are at node positions $\bm{q}$. In the figures throughout this paper, the nodes $\bm{q}$ are the vertices of the tetrahedral elements. The elements, which model the solid material, are shown in shades of blue. SOFA uses a formulation that accounts for the geometric nonlinearities of the deformation and the material is chosen to be characterized by Hooke's law (Young's modulus and Poisson's ratio).

During each step $t$ of the simulation, a linearization of the internal forces is computed:
\begin{equation}
\label{eq:linearization}
\mathbb{F}(\bm{q_{t}}) \approx \mathbb{F}(\bm{q_{t-1}}) + K(\bm{q_{t-1}})d\bm{q},
\end{equation}
where $d\bm{q = q_{t}-q_{t-1}}$ is the displacement of the nodes and $K=\frac{\partial \mathbb{F}(\bm{q_{t-1}})}{\partial \bm{q}}$ is the tangential stiffness matrix for the current node positions $\bm{q}$. The deformation is solved under the assumption of quasi-static motion. To complete the picture, external forces are included using Lagrangian multipliers:
\begin{equation}
\label{eq:equilibrium}
0 = -K(\bm{q_{t-1}})d\bm{q} + \mathbb{P} - \mathbb{F}(\bm{q_{t-1}}) + H^{T}\bm{\lambda}.
\end{equation}
$H^{T}\bm{\lambda}$ is a vector that gathers boundary forces, such as contacts or externally controlled inputs. The size of $\bm{\lambda}$ is equal to the number of rows in $H$ and to the number of actuators (contact forces, cables, etc.). $\mathbb{P}$ represents known external forces, such as gravity. Then, (\ref{eq:equilibrium}) is solved under the assumption of static equilibrium, producing a motion that is a succession of quasi-static states. Please refer to~\cite{duriez2013control} and \cite{coevoet2017software} for a more in-depth discussion about the FEM formulation used here.

\subsection{Barycentric Mapping}
\label{subsec:BarycentricMapping}
\begin{figure}
    \centering
    \small
    \includegraphics[width=0.8\linewidth]{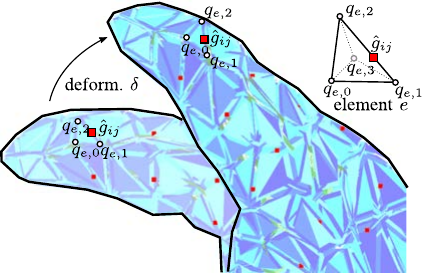}
    \caption{Position of a mapped grid point $\hat{g}_{ij}$ at rest and in a deformed configuration.}
    \label{fig:BarycentricMapping}
\end{figure}

In order to keep track of the movement of the grid-points $g_{ij}$ of the capacitive sensor or to apply a higher resolution visual model that follows the deformation, it becomes necessary to be able \emph{map} points to the underlying mechanical model. In SOFA, this mapping is provided by a component called \emph{Barycentric Mapping} (BM), illustrated by Fig.~\ref{fig:BarycentricMapping}. On the example of a grid-point $g_{ij}$ of the capacitive pad (see Sec.~\ref{subsubsec:WeightedPositionDetection}), that we wish to attach to the model as $\hat{g}_{ij}$, we see that there is one element $e$ that is closest to that point. Let $q_{e,0},q_{e,1},q_{e,2},q_{e,3}$ be the vertices of that element. Then, the coordinates of $\hat{g}_{ij}$ can be written as a weighted sum of the coordinates of the vertex nodes:
\begin{equation}
    \hat{g}_{ij} = a_{0}q_{e,0} + a_{1}q_{e,1} + a_{2}q_{e,2} + a_{3}q_{e,3}, \text{with} \sum_{u=0}^{3}a_{u}=1 
\end{equation}
During deformation, the coordinates of $\hat{g}_{ij}$ change, whereas the barycentric coordinates $b_{ij}=(a_{0},a_{1},a_{2},a_{3})$ remain constant throughout. These barycentric coordinates are used to update the coordinates of the mapped points as the robot deforms.

\subsection{Mapping 2D Touch Coordinates to 3D Touch Coordinates}
\label{subsec:Mapping2D_3D}

\begin{figure}
    \centering
    \small
    \includegraphics[width=0.79\linewidth]{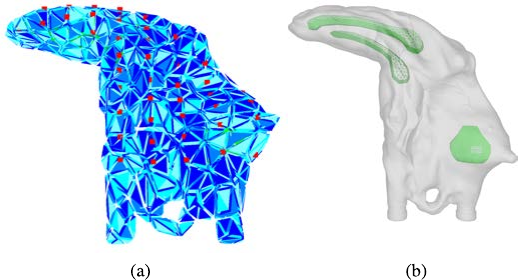}
    \caption{(a) In red, all the grid points $\hat{g}_{ij}$ mapped to the mechanical model. (b) The chambers embedded inside the Fruit that can be inflated to create an animation.}
    \label{fig:SimulationPoints+Cavities}
\end{figure}

We can now describe how the grid points on the Fruit are represented in the simulation. As discussed in Sec.~\ref{subsec:Integration}, the grid coordinates on the Fruit are calculated in preparation for the fabrication process. These coordinates can then be used to create barycentrically mapped points, that will move along when the Fruit deforms in simulation. All the mapped points of the grid are shown in Fig.~\ref{fig:SimulationPoints+Cavities}(a). During interaction, using the same weights $w_{ij}$ found for (\ref{eq:WeightedPosition}), we can calculate the weighted 3D coordinate $\hat{g}_{w}$ of the touch point as:
\begin{equation}
    \label{eq:Weighted3DPosition}
    \hat{g}_{w} = \sum_{(i,j)\in N_{9}}w_{ij}\hat{g}_{ij},
\end{equation}
where $\hat{g}_{ij}$ are now the 3D coordinates of the grid points mapped to the Fruit.

\subsection{Modeling of Pressure}
\label{subsec:PressureModling}
\begin{figure}
    \centering
    \small
    \includegraphics[width=0.6\linewidth]{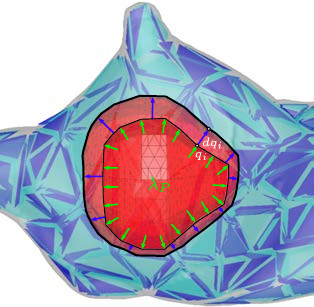}
    \caption{Pressure is represented as the effect of a Lagrangian multiplier $\lambda_{P}$. The corresponding force on each node $q_{i}$ leads to a displacement $dq_{i}$.}
    \label{fig:PressureModeling}
\end{figure}

The Lagrange multipliers in (\ref{eq:equilibrium}) can be used to model the effect of actuation due to the pressurization of chambers, as illustrated by Fig.~\ref{fig:PressureModeling}. A chamber is a hollow region inside the volumetric model delimited by a surface mesh that represents the enclosed volume. The chambers inside the fruit are shown in Fig.~\ref{fig:SimulationPoints+Cavities}(b). For each chamber, a variable $\lambda_{P}$ is used to represent the magnitude of the pressure $P$. This pressure is, by definition, distributed uniformly on the surface of the cavity, as shown by the green arrows in Fig.~\ref{fig:PressureModeling}. The exact force exerted on a node $q_{i}$ due to the pressure $P$ is proportional to the surface area of the triangles adjacent to this node on the surface mesh. These contributions are codified into the matrix $H^{T}$ in (\ref{eq:equilibrium}). The force on each node is responsible for a node motion $dq_{i}$ as the equation (\ref{eq:equilibrium}) for motion is solved. In the scope of this work, the pressure is created by syringe systems with moving pistons (see attached video). Pressure sensors included in the systems provide the measurements that are relayed to SOFA to update the model. For more details about this actuation model, please consult~\cite{coevoet2017optimization}.
 
\section{Experimental Validation}
\label{sec:Results}
To validate our proposed approach, we performed the following experiments: 1) We first verify in a \emph{sensitivity analysis} that the baseline capacitive value does not change significantly between a rest configuration and a deformed configuration. We test this with $4\!\times\!4$ flat touchpad and the Fruit. 2) We then verify the \emph{accuracy of position detection} both at rest and in a deformed configuration for the Fruit. These quantitative results are complemented by the attached video material.

\subsection{Sensitivity Analysis}
\subsubsection{$4\times4$ Touchpad}

\begin{figure}
    \centering
    \includegraphics[width=0.8\linewidth]
    {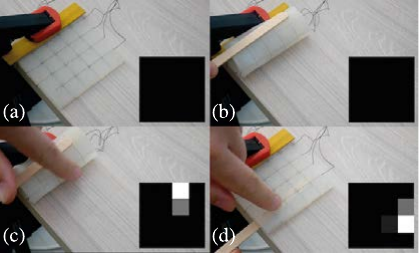}
    \caption{Deformation sequence for the $4\!\times\!4$ touchpad: (a) rest configuration, (b) $\SI{180}{\degree}$ bent configuration, (c) bent and touch from behind, and (d) rest and touch from front.}
    \label{fig:4x4Sensitivity}
\end{figure}

\begin{figure}
\centering
\includegraphics[width=0.63\linewidth]{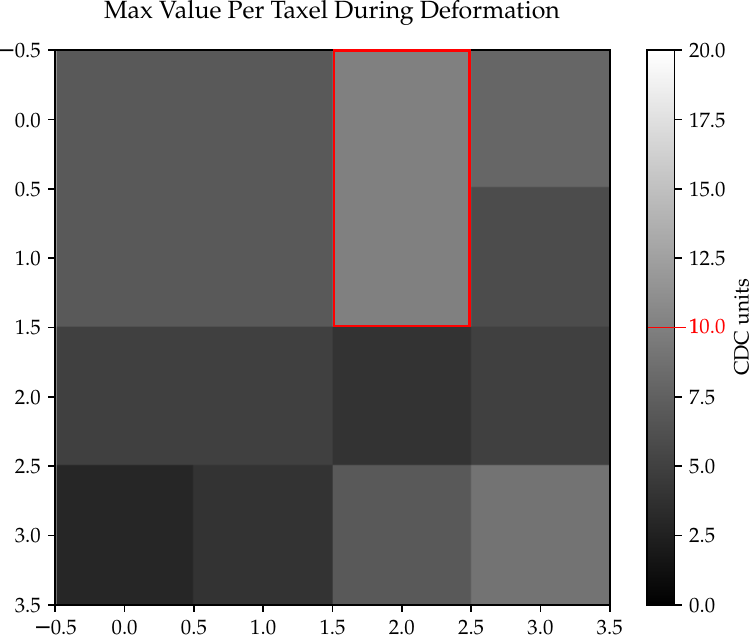}
\caption{Maximal shift in baseline value during the sequence Fig.~\ref{fig:4x4Sensitivity}(a)-(b). The highest value is 10 CDC units for two of the taxels.}
\label{fig:Max4x4}
\end{figure}

A flat touchpad implemented with the conductive yarn allows us to get a clear reference for the effect of deformations on the baseline capacitance value for each taxel. We have thus fabricated a flat $4\!\times\!4$ touchpad with dimensions $80\!\times80\!\times\SI{3}{\milli\meter\cubed}$, as seen in Fig.~\ref{fig:4x4Sensitivity}. The active area on this touchpad is $48\!\times\!\SI{48}{\milli\meter\squared}$, with a spacing between crossing points of $\SI{16}{\milli\meter}$. Throughout, we have established 20 CDC counts as the threshold for touch detection, which corresponds to very light touch or even pre-touch. The deformation sequence in Fig.~\ref{fig:4x4Sensitivity} shows the touchpad at rest in (a), then bent by in \SI{180}{\degree} in (b) where no taxel value surpasses the established threshold for touch. In (c), the configuration is the same as in (b), but an interaction with the human finger happens from the back side, triggering the touch readings. In (d) the touchpad is flat again and this time an interaction is registered from the front. Taking into account the side length of $\SI{48}{\milli\meter}$ of the active region of the touchpad, this means that each taxel rejects false touch detection with regards to bending with a curvature radius of at least $\frac{48}{\pi}=\SI{15.3}{\milli\meter}$ or less. This sequence is also a part of the media attachment. In Fig.~\ref{fig:Max4x4}, it is shown what the maximum deviation from the baseline capacitance at rest ($C_{0}$) for the bent configuration in Fig.~\ref{fig:4x4Sensitivity}~(b). It is shown that $C_{\delta}$ has a maximum difference of 10 CDC counts with respect to $C_{0}$, which is clearly below the established threshold for touch.
  
\subsubsection{The Fruit}

\begin{figure}
\centering
\includegraphics[width=0.63\linewidth]{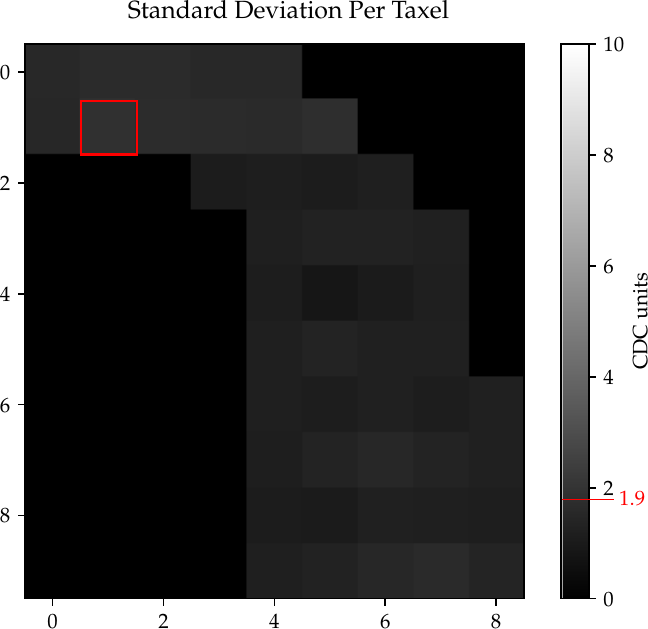}
\caption{Capacitive signal standard deviation of each taxel at rest. The highest values is 1.9 CDC units. The color scale covers half of the touch activation threshold (20 CDC units).}
\label{fig:Std}
\end{figure}

We measured the noise level for each of the taxels on the Fruit at rest in terms of the standard deviation from the average value for window of 300 frames, as shown by Fig.~\ref{fig:Std}. The maximal standard deviation for any taxel was 1.9 CDC units. Thus, the noise floor is significantly below the contact detection threshold of 20 CDC units. 

\begin{figure}
\centering
\includegraphics[width=0.6\linewidth]{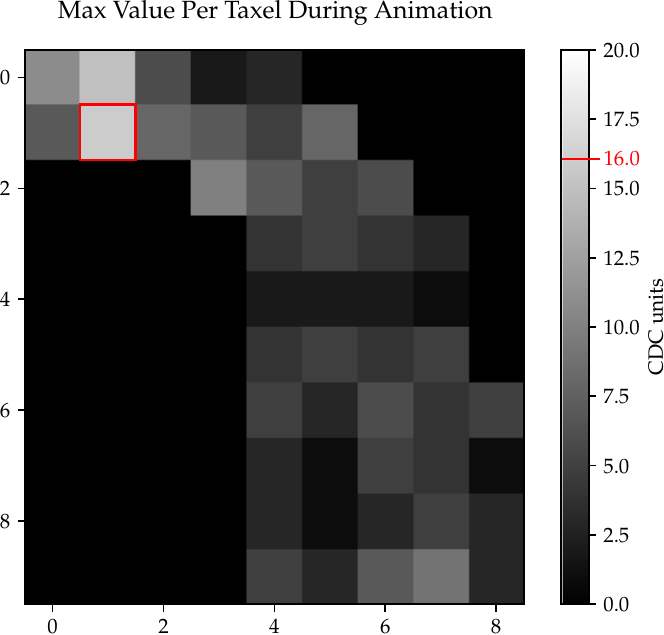}
\caption{Maximal shift in baseline value during the animation sequence designed for the art installation ($C_{\delta}$). The highest value is 16 CDC units.}
\label{fig:Max}
\end{figure}

To quantify the effect of deformation on the baseline value, we recorded the maximum shift in baseline value for each taxel over the course of the animation designed for the art installation, where the three embedded chambers are inflated and deflated. The results are shown in Fig.~\ref{fig:Max}. The maximal shift is detected at 16 CDC units and on average, over the whole grid, the maximum shift is 5.0 CDC units. We observe that some taxels at the tip exhibit significantly larger shifts than the rest of the shape. However, there is no significant difference between the maximal deformation in the different regions of the Fruit. Therefore, we assume that the difference in baseline shift is due to issues related to fabrication, e.\,g.\ crossing points that deforms more than we intended. By revisiting the fabrication process, we should be able to reduce this effect. Even at the reported 16 CDC units, the shift is below the touch detection threshold.  

\subsection{Accuracy of Position Detection}
\begin{figure}
    \centering
    \includegraphics[width=1\linewidth]
    {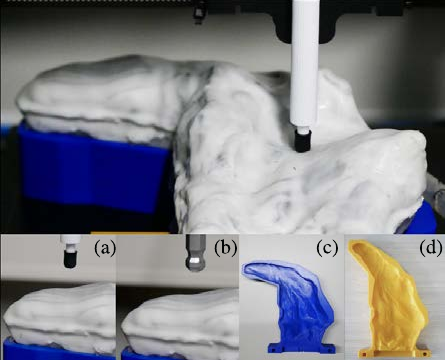}
    \caption{The experimental setup for testing position detection accuracy consists of a modified 3D printer and a jig keeping the Fruit at a defined position. (a) Shows the small conductive rubber tip indenter. (b) shows the medium-sized indenter with an Allen key. Both are grounded and provide proper detection by the capacitive sensor. (c) Shows the jig for the Fruit in its rest position. (d) Shows the jig for the Fruit in its deformed configuration.}
    \label{fig:ExperimentSetup}
\end{figure}

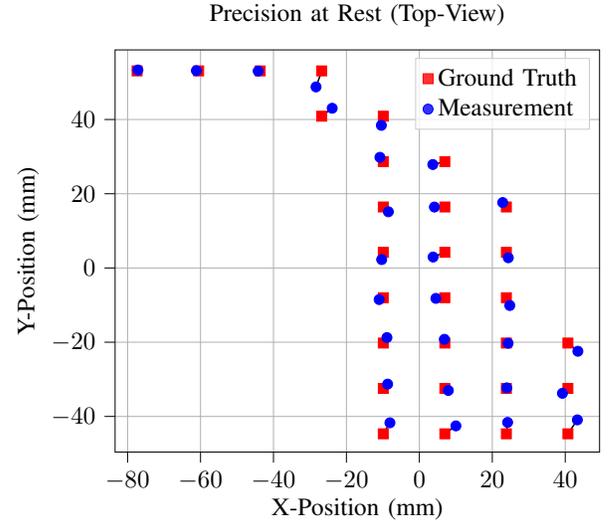
\begin{figure}
\centering
 \resizebox{0.9\linewidth}{!}{%
\begin{tikzpicture}

\definecolor{darkgray176}{RGB}{176,176,176}
\definecolor{lightgray204}{RGB}{204,204,204}

\begin{axis}[
title={Precision at Rest (Top-View)},
legend cell align={left},
legend style={fill opacity=0.8, draw opacity=1, text opacity=1, draw=lightgray204},
tick align=outside,
tick pos=left,
x grid style={darkgray176},
xlabel={X-Position (mm)},
xmajorgrids,
xmin=-83.4306612815448, xmax=49.4145765676138,
xtick style={color=black},
y grid style={darkgray176},
ylabel={Y-Position (mm)},
ymajorgrids,
ymin=-49.5994240574957, ymax=58.7526561652635,
ytick style={color=black}
]
\addplot [draw=red, fill=red, mark=square*, only marks]
table{%
x  y
-77.3922413793103 53.1034482758621
-60.5172413793103 53.1034482758621
-43.6422413793103 53.1034482758621
-26.7672413793103 40.8812260536398
-26.7672413793103 53.1034482758621
-9.89224137931035 -44.6743295019157
-9.89224137931035 -32.4521072796935
-9.89224137931035 -20.2298850574713
-9.89224137931035 -8.00766283524902
-9.89224137931035 4.21455938697319
-9.89224137931035 16.4367816091954
-9.89224137931035 28.6590038314176
-9.89224137931035 40.8812260536398
6.98275862068965 -44.6743295019157
6.98275862068965 -32.4521072796935
6.98275862068965 -20.2298850574713
6.98275862068965 -8.00766283524902
6.98275862068965 4.21455938697319
6.98275862068965 16.4367816091954
6.98275862068965 28.6590038314176
23.8577586206897 -44.6743295019157
23.8577586206897 -32.4521072796935
23.8577586206897 -20.2298850574713
23.8577586206897 -8.00766283524902
23.8577586206897 4.21455938697319
23.8577586206897 16.4367816091954
40.7327586206897 -44.6743295019157
40.7327586206897 -32.4521072796935
40.7327586206897 -20.2298850574713
};
\addlegendentry{Ground Truth}
\addplot [draw=blue, fill=blue, mark=*, only marks]
table{%
x  y
-77.1736542994332 53.3370383592779
-61.1357078708618 53.1672852728581
-44.1976368575727 53.0606961255713
-23.9012071163346 43.0323087470365
-28.3299042994332 48.7876556432285
-8.06793621432678 -41.7174665001915
-8.70267215657603 -31.2970004414276
-8.89240429943316 -18.7514390069773
-11.0247572406096 -8.51299795154276
-10.3299042994332 2.26172971730256
-8.50177929943317 15.1425939148334
-10.812047156576 29.8141106696835
-10.4211205156494 38.4262942818671
10.0193604064492 -42.5434990408674
7.94227311992169 -32.9960958502004
6.8665242719954 -19.220280864708
4.53616712913827 -8.19118033560749
3.76888602314748 2.94775122267891
4.13178924895393 16.4184800159883
3.68572070056683 27.8740753963149
24.1974833410163 -41.6064350767049
23.9690414836994 -32.2962354768072
24.3287495467207 -20.2835694279966
24.8007775187486 -10.1065242509514
24.3843814148525 2.75061860619145
22.8469824930197 17.6248324426694
43.322070932074 -40.9211830425652
39.2249737493473 -33.7669965699603
43.4655502460214 -22.4345665789937
};
\addlegendentry{Measurement}

\addplot [semithick, black, forget plot]
table {%
-77.1736542994332 53.3370383592779
-77.3922413793103 53.1034482758621
};
\addplot [semithick, black, forget plot]
table {%
-61.1357078708618 53.1672852728581
-60.5172413793103 53.1034482758621
};
\addplot [semithick, black, forget plot]
table {%
-44.1976368575727 53.0606961255713
-43.6422413793103 53.1034482758621
};
\addplot [semithick, black, forget plot]
table {%
-23.9012071163346 43.0323087470365
-26.7672413793103 40.8812260536398
};
\addplot [semithick, black, forget plot]
table {%
-28.3299042994332 48.7876556432285
-26.7672413793103 53.1034482758621
};
\addplot [semithick, black, forget plot]
table {%
-8.06793621432678 -41.7174665001915
-9.89224137931035 -44.6743295019157
};
\addplot [semithick, black, forget plot]
table {%
-8.70267215657603 -31.2970004414276
-9.89224137931035 -32.4521072796935
};
\addplot [semithick, black, forget plot]
table {%
-8.89240429943316 -18.7514390069773
-9.89224137931035 -20.2298850574713
};
\addplot [semithick, black, forget plot]
table {%
-11.0247572406096 -8.51299795154276
-9.89224137931035 -8.00766283524902
};
\addplot [semithick, black, forget plot]
table {%
-10.3299042994332 2.26172971730256
-9.89224137931035 4.21455938697319
};
\addplot [semithick, black, forget plot]
table {%
-8.50177929943317 15.1425939148334
-9.89224137931035 16.4367816091954
};
\addplot [semithick, black, forget plot]
table {%
-10.812047156576 29.8141106696835
-9.89224137931035 28.6590038314176
};
\addplot [semithick, black, forget plot]
table {%
-10.4211205156494 38.4262942818671
-9.89224137931035 40.8812260536398
};
\addplot [semithick, black, forget plot]
table {%
10.0193604064492 -42.5434990408674
6.98275862068965 -44.6743295019157
};
\addplot [semithick, black, forget plot]
table {%
7.94227311992169 -32.9960958502004
6.98275862068965 -32.4521072796935
};
\addplot [semithick, black, forget plot]
table {%
6.8665242719954 -19.220280864708
6.98275862068965 -20.2298850574713
};
\addplot [semithick, black, forget plot]
table {%
4.53616712913827 -8.19118033560749
6.98275862068965 -8.00766283524902
};
\addplot [semithick, black, forget plot]
table {%
3.76888602314748 2.94775122267891
6.98275862068965 4.21455938697319
};
\addplot [semithick, black, forget plot]
table {%
4.13178924895393 16.4184800159883
6.98275862068965 16.4367816091954
};
\addplot [semithick, black, forget plot]
table {%
3.68572070056683 27.8740753963149
6.98275862068965 28.6590038314176
};
\addplot [semithick, black, forget plot]
table {%
24.1974833410163 -41.6064350767049
23.8577586206897 -44.6743295019157
};
\addplot [semithick, black, forget plot]
table {%
23.9690414836994 -32.2962354768072
23.8577586206897 -32.4521072796935
};
\addplot [semithick, black, forget plot]
table {%
24.3287495467207 -20.2835694279966
23.8577586206897 -20.2298850574713
};
\addplot [semithick, black, forget plot]
table {%
24.8007775187486 -10.1065242509514
23.8577586206897 -8.00766283524902
};
\addplot [semithick, black, forget plot]
table {%
24.3843814148525 2.75061860619145
23.8577586206897 4.21455938697319
};
\addplot [semithick, black, forget plot]
table {%
22.8469824930197 17.6248324426694
23.8577586206897 16.4367816091954
};
\addplot [semithick, black, forget plot]
table {%
43.322070932074 -40.9211830425652
40.7327586206897 -44.6743295019157
};
\addplot [semithick, black, forget plot]
table {%
39.2249737493473 -33.7669965699603
40.7327586206897 -32.4521072796935
};
\addplot [semithick, black, forget plot]
table {%
43.4655502460214 -22.4345665789937
40.7327586206897 -20.2298850574713
};
\end{axis}

\end{tikzpicture}}
\caption{Top-view of the measured vs.\ ground truth positions on the sampled grid at rest.}
\label{fig:EvalFruitRestXY}
\end{figure}

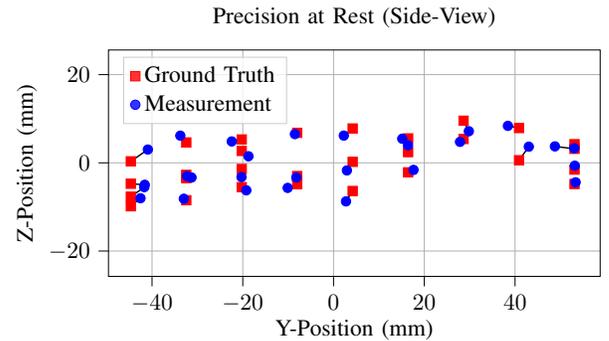
\begin{figure}
\centering
 \resizebox{0.9\linewidth}{!}{%
\begin{tikzpicture}

\definecolor{darkgray176}{RGB}{176,176,176}
\definecolor{lightgray204}{RGB}{204,204,204}

\begin{axis}[
title={Precision at Rest (Side-View)},
legend cell align={left},
legend style={
  fill opacity=0.8,
  draw opacity=1,
  text opacity=1,
  at={(0.03,0.97)},
  anchor=north west,
  draw=lightgray204
},
height=5cm,
width=9cm,
tick align=outside,
tick pos=left,
x grid style={darkgray176},
xlabel={Y-Position (mm)},
xmajorgrids,
xmin=-49.5994240574957, xmax=58.7526561652635,
xtick style={color=black},
y grid style={darkgray176},
ylabel={Z-Position (mm)},
ymajorgrids,
ymin=-25.7467914177829, ymax=25.5872931546899,
ytick style={color=black}
]
\addplot [draw=red, fill=red, mark=square*, only marks]
table{%
x  y
53.1034482758621 -4.79420760845201
53.1034482758621 -1.54303976439226
53.1034482758621 4.22107366962638
40.8812260536398 0.589280169694462
53.1034482758621 3.17406262745965
-44.6743295019157 -7.68126667160797
-32.4521072796935 -3.45473239806122
-20.2298850574713 2.69311157218316
-8.00766283524902 6.80243509188628
4.21455938697319 7.79186930489919
16.4367816091954 5.54441267251812
28.6590038314176 9.54859652051798
40.8812260536398 7.88806980874821
-44.6743295019157 -9.84602581809365
-32.4521072796935 -8.47998417385529
-20.2298850574713 -5.5133350037944
-8.00766283524902 -3.00325969927835
4.21455938697319 0.223690600195496
16.4367816091954 2.43718945055068
28.6590038314176 5.40515629969875
-44.6743295019157 -4.7413220206539
-32.4521072796935 -2.7395056755321
-20.2298850574713 -1.40496469281143
-8.00766283524902 -4.84059800225253
4.21455938697319 -6.39087905783603
16.4367816091954 -2.15211050439166
-44.6743295019157 0.33834207096254
-32.4521072796935 4.62927606889534
-20.2298850574713 5.29866516317645
};
\addlegendentry{Ground Truth}
\addplot [draw=blue, fill=blue, mark=*, only marks]
table{%
x  y
53.3370383592779 -4.41364829575735
53.1672852728581 -0.658971240572122
53.0606961255713 3.30422546051591
43.0323087470365 3.65600346381672
48.7876556432285 3.72618865145374
-41.7174665001915 -5.5156838863095
-31.2970004414276 -3.32597673998211
-18.7514390069773 1.49001947971655
-8.51299795154276 6.51476688097657
2.26172971730256 6.17360416841364
15.1425939148334 5.46233020479143
29.8141106696835 7.15937590760477
38.4262942818671 8.38146149138147
-42.5434990408674 -8.02440067658192
-32.9960958502004 -8.14479344609937
-19.220280864708 -6.23308815161134
-8.19118033560749 -3.32828854393423
2.94775122267891 -1.72347339904137
16.4184800159883 3.9855778233447
27.8740753963149 4.75511669907609
-41.6064350767049 -4.98105495408315
-32.2962354768072 -3.02472420640411
-20.2835694279966 -3.23127778424863
-10.1065242509514 -5.68435749146024
2.75061860619145 -8.73989479176252
17.6248324426694 -1.60642130953736
-40.9211830425652 3.01264032431856
-33.7669965699603 6.16128816739271
-22.4345665789937 4.85345619458231
};
\addlegendentry{Measurement}
\addplot [semithick, black, forget plot]
table {%
53.3370383592779 -4.41364829575735
53.1034482758621 -4.79420760845201
};
\addplot [semithick, black, forget plot]
table {%
53.1672852728581 -0.658971240572122
53.1034482758621 -1.54303976439226
};
\addplot [semithick, black, forget plot]
table {%
53.0606961255713 3.30422546051591
53.1034482758621 4.22107366962638
};
\addplot [semithick, black, forget plot]
table {%
43.0323087470365 3.65600346381672
40.8812260536398 0.589280169694462
};
\addplot [semithick, black, forget plot]
table {%
48.7876556432285 3.72618865145374
53.1034482758621 3.17406262745965
};
\addplot [semithick, black, forget plot]
table {%
-41.7174665001915 -5.5156838863095
-44.6743295019157 -7.68126667160797
};
\addplot [semithick, black, forget plot]
table {%
-31.2970004414276 -3.32597673998211
-32.4521072796935 -3.45473239806122
};
\addplot [semithick, black, forget plot]
table {%
-18.7514390069773 1.49001947971655
-20.2298850574713 2.69311157218316
};
\addplot [semithick, black, forget plot]
table {%
-8.51299795154276 6.51476688097657
-8.00766283524902 6.80243509188628
};
\addplot [semithick, black, forget plot]
table {%
2.26172971730256 6.17360416841364
4.21455938697319 7.79186930489919
};
\addplot [semithick, black, forget plot]
table {%
15.1425939148334 5.46233020479143
16.4367816091954 5.54441267251812
};
\addplot [semithick, black, forget plot]
table {%
29.8141106696835 7.15937590760477
28.6590038314176 9.54859652051798
};
\addplot [semithick, black, forget plot]
table {%
38.4262942818671 8.38146149138147
40.8812260536398 7.88806980874821
};
\addplot [semithick, black, forget plot]
table {%
-42.5434990408674 -8.02440067658192
-44.6743295019157 -9.84602581809365
};
\addplot [semithick, black, forget plot]
table {%
-32.9960958502004 -8.14479344609937
-32.4521072796935 -8.47998417385529
};
\addplot [semithick, black, forget plot]
table {%
-19.220280864708 -6.23308815161134
-20.2298850574713 -5.5133350037944
};
\addplot [semithick, black, forget plot]
table {%
-8.19118033560749 -3.32828854393423
-8.00766283524902 -3.00325969927835
};
\addplot [semithick, black, forget plot]
table {%
2.94775122267891 -1.72347339904137
4.21455938697319 0.223690600195496
};
\addplot [semithick, black, forget plot]
table {%
16.4184800159883 3.9855778233447
16.4367816091954 2.43718945055068
};
\addplot [semithick, black, forget plot]
table {%
27.8740753963149 4.75511669907609
28.6590038314176 5.40515629969875
};
\addplot [semithick, black, forget plot]
table {%
-41.6064350767049 -4.98105495408315
-44.6743295019157 -4.7413220206539
};
\addplot [semithick, black, forget plot]
table {%
-32.2962354768072 -3.02472420640411
-32.4521072796935 -2.7395056755321
};
\addplot [semithick, black, forget plot]
table {%
-20.2835694279966 -3.23127778424863
-20.2298850574713 -1.40496469281143
};
\addplot [semithick, black, forget plot]
table {%
-10.1065242509514 -5.68435749146024
-8.00766283524902 -4.84059800225253
};
\addplot [semithick, black, forget plot]
table {%
2.75061860619145 -8.73989479176252
4.21455938697319 -6.39087905783603
};
\addplot [semithick, black, forget plot]
table {%
17.6248324426694 -1.60642130953736
16.4367816091954 -2.15211050439166
};
\addplot [semithick, black, forget plot]
table {%
-40.9211830425652 3.01264032431856
-44.6743295019157 0.33834207096254
};
\addplot [semithick, black, forget plot]
table {%
-33.7669965699603 6.16128816739271
-32.4521072796935 4.62927606889534
};
\addplot [semithick, black, forget plot]
table {%
-22.4345665789937 4.85345619458231
-20.2298850574713 5.29866516317645
};
\end{axis}

\end{tikzpicture}}
\caption{Side-view of the measured vs.\ ground truth positions on the sampled grid at rest.}
\label{fig:EvalFruitRestYZ}
\end{figure}

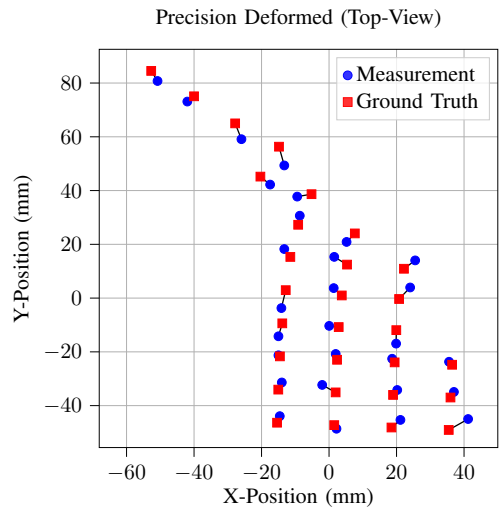
\begin{figure}
\centering
 \resizebox{0.75\linewidth}{!}{%
\begin{tikzpicture}

\definecolor{darkgray176}{RGB}{176,176,176}
\definecolor{lightgray204}{RGB}{204,204,204}

\begin{axis}[
title={Precision Deformed (Top-View)},
legend cell align={left},
legend style={fill opacity=0.8, draw opacity=1, text opacity=1, draw=lightgray204},
height=8cm,
width=8cm,
tick align=outside,
tick pos=left,
x grid style={darkgray176},
xlabel={X-Position (mm)},
xmajorgrids,
xmin=-68.0908467171512, xmax=50,
xtick style={color=black},
y grid style={darkgray176},
ylabel={Y-Position (mm)},
ymajorgrids,
ymin=-55.6, ymax=92.5464427930894,
ytick distance=20,
ytick style={color=black}
]

\addplot [draw=blue, fill=blue, mark=*, only marks]
table{%
x  y
-50.8193966918718 80.7599178188833
-41.9944253124681 73.1086889344884
-25.9582681681947 59.1083971042192
-17.4572617201159 42.224523336313
-13.2632032077712 49.341092868547
-14.5983264254901 -43.9146335963364
-13.9962006482449 -31.3867518899867
-14.9913975368386 -21.2763684668169
-14.9892932343489 -14.2077353944215
-14.1220028251643 -3.7158619734958
-13.2642555217148 18.2409047887768
-8.6697055411491 30.649390861891
-9.43073485282734 37.7516408597575
2.20295671206206 -48.6027857320078
-2.02673811089951 -32.3026863605549
1.97214554617484 -20.7835040159183
0.0521729705798037 -10.3327002176742
1.38294528431573 3.71634882837143
1.53790281496061 15.3310198238034
5.19467918787436 20.8897337131926
21.1514749882552 -45.3149918458051
20.2039027628629 -34.1892815535655
18.6959035171679 -22.5631010437824
19.8688567669137 -16.9341862421201
24.0321658856972 3.94929511335424
25.5203645119538 14.0263967114066
41.2064791834092 -44.98420270868
37.0010638948027 -34.9030210099351
35.558195770069 -23.6855387119037
};
\addlegendentry{Measurement}
\addplot [draw=red, fill=red, mark=square*, only marks]
table{%
x  y
-52.6933703717747 84.5112356058286
-39.9985469536526 75.0653281445805
-27.8209500226273 64.9944870316574
-20.3415010183282 45.1924217511704
-14.8301443394591 56.3210733035869
-15.3996094155184 -46.3534923899392
-15.0405898745791 -34.0453567606887
-14.5591974483541 -21.6786526310426
-13.8546972091369 -9.3940902260649
-12.8308694270597 2.9506160483101
-11.4726472785422 15.3109511824678
-9.14527749112138 27.2764271950624
-5.19172374765914 38.6537518696137
1.53587937881532 -47.2593086810306
1.95446538469972 -35.0795875209562
2.37530490467316 -22.9429359134628
2.86266515825227 -10.7785490199569
3.77296365074706 0.98246326375191
5.32821025034708 12.4596606730187
7.63890439454012 24.0595289651137
18.4929752863561 -48.138147103242
18.9721133562196 -36.0165251682441
19.439272833743 -23.9115271018435
19.9264228270061 -11.9239233924772
20.7607452548144 -0.33614108060263
22.1969086100024 10.9133937244852
35.4668568437261 -49.0493953411986
35.9830771283849 -36.9844279589322
36.4723593354851 -24.7992784689654
};
\addlegendentry{Ground Truth}

\addplot [semithick, black, forget plot]
table {%
-50.8193966918718 80.7599178188833
-52.6933703717747 84.5112356058286
};
\addplot [semithick, black, forget plot]
table {%
-41.9944253124681 73.1086889344884
-39.9985469536526 75.0653281445805
};
\addplot [semithick, black, forget plot]
table {%
-25.9582681681947 59.1083971042192
-27.8209500226273 64.9944870316574
};
\addplot [semithick, black, forget plot]
table {%
-17.4572617201159 42.224523336313
-20.3415010183282 45.1924217511704
};
\addplot [semithick, black, forget plot]
table {%
-13.2632032077712 49.341092868547
-14.8301443394591 56.3210733035869
};
\addplot [semithick, black, forget plot]
table {%
-14.5983264254901 -43.9146335963364
-15.3996094155184 -46.3534923899392
};
\addplot [semithick, black, forget plot]
table {%
-13.9962006482449 -31.3867518899867
-15.0405898745791 -34.0453567606887
};
\addplot [semithick, black, forget plot]
table {%
-14.9913975368386 -21.2763684668169
-14.5591974483541 -21.6786526310426
};
\addplot [semithick, black, forget plot]
table {%
-14.9892932343489 -14.2077353944215
-13.8546972091369 -9.3940902260649
};
\addplot [semithick, black, forget plot]
table {%
-14.1220028251643 -3.7158619734958
-12.8308694270597 2.9506160483101
};
\addplot [semithick, black, forget plot]
table {%
-13.2642555217148 18.2409047887768
-11.4726472785422 15.3109511824678
};
\addplot [semithick, black, forget plot]
table {%
-8.6697055411491 30.649390861891
-9.14527749112138 27.2764271950624
};
\addplot [semithick, black, forget plot]
table {%
-9.43073485282734 37.7516408597575
-5.19172374765914 38.6537518696137
};
\addplot [semithick, black, forget plot]
table {%
2.20295671206206 -48.6027857320078
1.53587937881532 -47.2593086810306
};
\addplot [semithick, black, forget plot]
table {%
-2.02673811089951 -32.3026863605549
1.95446538469972 -35.0795875209562
};
\addplot [semithick, black, forget plot]
table {%
1.97214554617484 -20.7835040159183
2.37530490467316 -22.9429359134628
};
\addplot [semithick, black, forget plot]
table {%
0.0521729705798037 -10.3327002176742
2.86266515825227 -10.7785490199569
};
\addplot [semithick, black, forget plot]
table {%
1.38294528431573 3.71634882837143
3.77296365074706 0.98246326375191
};
\addplot [semithick, black, forget plot]
table {%
1.53790281496061 15.3310198238034
5.32821025034708 12.4596606730187
};
\addplot [semithick, black, forget plot]
table {%
5.19467918787436 20.8897337131926
7.63890439454012 24.0595289651137
};
\addplot [semithick, black, forget plot]
table {%
21.1514749882552 -45.3149918458051
18.4929752863561 -48.138147103242
};
\addplot [semithick, black, forget plot]
table {%
20.2039027628629 -34.1892815535655
18.9721133562196 -36.0165251682441
};
\addplot [semithick, black, forget plot]
table {%
18.6959035171679 -22.5631010437824
19.439272833743 -23.9115271018435
};
\addplot [semithick, black, forget plot]
table {%
19.8688567669137 -16.9341862421201
19.9264228270061 -11.9239233924772
};
\addplot [semithick, black, forget plot]
table {%
24.0321658856972 3.94929511335424
20.7607452548144 -0.33614108060263
};
\addplot [semithick, black, forget plot]
table {%
25.5203645119538 14.0263967114066
22.1969086100024 10.9133937244852
};
\addplot [semithick, black, forget plot]
table {%
41.2064791834092 -44.98420270868
35.4668568437261 -49.0493953411986
};
\addplot [semithick, black, forget plot]
table {%
37.0010638948027 -34.9030210099351
35.9830771283849 -36.9844279589322
};
\addplot [semithick, black, forget plot]
table {%
35.558195770069 -23.6855387119037
36.4723593354851 -24.7992784689654
};
\end{axis}

\end{tikzpicture}}
\caption{Top-view of the measured vs.\ ground truth positions on the sampled grid under deformation.}
\label{fig:EvalFruitDeformedXY}
\end{figure}

To test the accuracy of position detection, we implemented a test bench, as shown in Fig.~\ref{fig:ExperimentSetup}. The test bench is a modified Artillery Hornet 3D-printer, where we replaced the extruder with the attachment piece for the indenters. The printer has a positioning accuracy of $\SI{0.1}{\milli\meter}$. Two jigs were designed to establish the experimental conditions. The first jig, shown in Fig.~\ref{fig:ExperimentSetup}~(c), holds the Fruit in its rest position, while the second jig, shown in Fig.~\ref{fig:ExperimentSetup}~(d), holds the Fruit in a deformed configuration. Fig.~\ref{fig:BarycentricMapping} shows a detail in simulation of the same deformed state. A series of sampling points on the Fruit are established near the capacitive grid points. We did not sample directly \emph{at} the grid points, as this biased the detection to the coordinates of that particular grid point. Instead, an offset of a quarter taxel size was applied to ensure that more than one taxel was activated by the presence of the indenter. The ground truth positions are shown in red in Figs.~\ref{fig:EvalFruitRestXY}, \ref{fig:EvalFruitRestYZ} and~\ref{fig:EvalFruitDeformedXY}. The blue points represent the positions detected by using (\ref{eq:Weighted3DPosition}). The sampling is carried out within our test-bench by approaching the Fruit from above until the contact threshold is reached (see also attached video). The deformed configuration is created by a rotation of the tip points of $\SI{30}{\degree}$. In simulation, this effect is achieved by springs and in the experiments by the jig of the deformed configuration. 

The summary of the results are provided in Table~\ref{tab:FruitResults}, where we report the average error and the standard deviation in $\si{mm}$ and $\si{\percent}$ for the different experimental conditions. The percentage error is expressed in relation to the largest distance between two sampling points (top-left to lower right in Fig.~\ref{fig:EvalFruitRestXY}, i.\,e. $\SI{152}{\milli\meter}$). The best results are achieved with the small indenter. At rest, the average error is $\SI{2.6}{\milli\meter}$ with standard deviation of $\SI{1.3}{\milli\meter}$. Slightly worse results are obtained for the medium indenter. This results show the potential for accurate contact location on the whole surface. As seen in Fig.~\ref{fig:EvalFruitRestXY}, there are no larger systematic biases for the contact location. However, we think that approaching the surface along the normal direction at the sampling point would prevent some smaller bias present in the results. For the deformed configuration, the jig shown in Fig.~\ref{fig:ExperimentSetup}~(d) was used. In this experiment, the average error is $\SI{4.4}{\milli\meter}$ with standard deviation of $\SI{1.8}{\milli\meter}$. While the precision is somewhat lower compared to the configuration at rest, this result confirms that the approach is capable of detecting contact positions satisfactorily, even when the shape is significantly deformed. For further studies, it would be interesting to increase the spatial resolution of the grid, which should improve the detection accuracy. This is possible with the current electronics and will be considered for future designs.

\begin{table}
\centering
\caption{Error of Position Detection for the Fruit}
\begin{tabular}{||c|c|c|c||} 
 \hline
 Configuration & Indenter & Avg. Error  & Error Std \\ [0.5ex]
 \hline\hline
 
 Rest & Small  & 2.6\si{\milli\meter(1.7\%)} & 1.3\si{\milli\meter} (0.8\%) \\
 \hline
  Rest & Medium  & 3.3\si{\milli\meter(2.2\%)} & 1.7\si{\milli\meter} (1.1\%) \\
 \hline 

 Deformed & Small  &4.4\si{\milli\meter} (2.8\%) & 1.8\si{\milli\meter} (1.1\%) \\
 \hline
 
\end{tabular}

\label{tab:FruitResults}
\end{table}

\section{Conclusions}
\label{sec:Conclusions}

In this paper, we have presented a capacitive touch technology for HRI in soft robotics. Our motivation is to provide a performant touch technology, which is essential for deploying relevant touch-based applications in soft HRI. This work has two main focuses: 1) Describing and validating that the touch technology is selectively sensitive to human touch, while it is robust against phantom detections due to underlying deformations of the robot. 2) Describing and validating how a solid mechanics model can be used to represent correct 3D touch position, taking into account the arbitrary shape of the object as well as the deformation caused by actuation. We evaluated the accuracy of position detection and robustness towards underlying deformations. As an application, we concentrated on demonstrating the potential this has for the Arts. In collaboration with a visual artist, we have created a touch-interactive soft robotics sculpture. The sculpture allows the user to explore the shape of the object by touch and also feel the respiratory movements it exhibits. This is considered to be an innovative approach, because typically art pieces are not designed to be touched and even less often they feature touch as the leading interaction modality. 

Further application domains of this technology are phantoms for medical training or for the study of surgical interventions. Another domain, where physical twins have been proposed, is agriculture. Recreating sensorized fruits can help improve many aspects worth automating, for example, harvesting or packaging \cite{Junge2022harvesting}. Additionally, for the neuroscientific and neurophysiological studies of humans through touch, with applications in affective touch within Human-Robot Interaction \cite{vyas2023affective}. Our ongoing efforts are in modeling further promising sensing principles for these types of applications, such as magnetic or resistive. Overall, we aim at modeling both the forces/deformations due to actuation as well as due to interactions for these devices, in order to have the complete picture. 

\renewcommand{\thetable}{\arabic{table}}

\addtolength{\textheight}{-7cm}   


\bibliographystyle{IEEEtran}

\bibliography{Bibliography.bib}

\end{document}